\documentclass[runningheads]{llncs}
\usepackage[T1]{fontenc}
\usepackage{graphicx,verbatim}
\usepackage{amsmath}
\usepackage{amsfonts}
\usepackage{tabularx,booktabs}
\usepackage{multirow}
\usepackage{placeins}
\usepackage{subcaption}
\usepackage{cite}
\usepackage{hyperref}
\newcolumntype{C}{>{\centering\arraybackslash}X} 
\usepackage{lipsum} 
\usepackage{pifont}
\begin{document}
\title{CellStyle: Improved Zero-Shot Cell Segmentation via Style Transfer}

\author{Rüveyda Yilmaz (\ding{41})\orcidID{0009-0007-7351-698X} \and
Zhu Chen\orcidID{0009-0009-9847-7686} \and \\ 
Yuli Wu\orcidID{0000-0002-6216-4911} \and
Johannes Stegmaier\orcidID{0000-0003-4072-3759}}
%
\authorrunning{R. Yilmaz et al.}
%
\institute{Institute of Imaging and Computer Vision, RWTH Aachen University, Germany \\Email:
    \email{rueveyda.yilmaz@lfb.rwth-aachen.de}}
    
\maketitle              
\begin{abstract}
Cell microscopy data are abundant; however, corresponding segmentation annotations remain scarce.
Moreover, variations in cell types, imaging devices, and staining techniques introduce significant domain gaps between datasets. 
As a result, even large, pretrained segmentation models trained on diverse datasets (source datasets) struggle to generalize to unseen datasets (target datasets).
To overcome this generalization problem, we propose CellStyle, which improves the segmentation quality of such models without requiring labels for the target dataset, thereby enabling zero-shot adaptation.
CellStyle transfers the attributes of an unannotated target dataset, such as texture, color, and noise, to the annotated source dataset.
This transfer is performed while preserving the cell shapes of the source images, ensuring that the existing source annotations can still be used while maintaining the visual characteristics of the target dataset.
The styled synthetic images with the existing annotations enable the finetuning of a generalist segmentation model for application to the unannotated target data. 
We demonstrate that CellStyle significantly improves the cell segmentation performance across diverse datasets by finetuning multiple segmentation models on the style-transferred data. 
The source code for CellStyle is publicly available at \href{https://github.com/ruveydayilmaz0/cellStyle}{https://github.com/ruveydayilmaz0/cellStyle}.

\end{abstract}
\section{Introduction}
Automated cell segmentation is essential for biomedical research, facilitating the extraction and analysis of cellular morphology and spatial organization \cite{durkee2021artificial}.
However, achieving accurate instance segmentation remains challenging due to the substantial variability in imaging modalities, cell types, and staining protocols \cite{gogoberidze2024defining,stacke2020measuring}.
This is caused by domain shifts, leading to failed generalization when applied to unseen microscopy datasets with different characteristics \cite{keaton2023celltranspose}.
To address this, previous studies \cite{cellpose,mediar,mesmer} have proposed generalist models trained on diverse datasets to enhance applicability across different datasets.
Often, these generalist models are finetuned on the target dataset by a human-in-the-loop approach \cite{cellpose} or through pre-generated labels \cite{mediar,stardist,cellpose}.
Alternatively, other studies \cite{sturm2024syncellfactory,eschweiler2024denoising,yilmaz2024annotated,yilmaz2024cascaded,svoboda2016mitogen,bahr2021cellcyclegan} have addressed the challenge of cell instance segmentation using classical methods, diffusion models or generative adversarial networks (GANs) by generating annotated synthetic images derived from a limited set of labeled real data or
by domain transfer, which adapts the visual features from one dataset to another one \cite{keaton2023celltranspose,yang2021minimizing,haq2020adversarial,liu2020unsupervised,bouteldja2022tackling}.
The generated images are then utilized as training datasets for instance segmentation models.
Although there are improvements in the segmentation quality, these models require supervised \cite{sturm2024syncellfactory,eschweiler2024denoising,yilmaz2024annotated,yilmaz2024cascaded,bahr2021cellcyclegan,keaton2023celltranspose,bouteldja2022tackling} or unsupervised \cite{yang2021minimizing,haq2020adversarial,liu2020unsupervised} training of a generative model or manual adjustments to the simulated image models \cite{svoboda2016mitogen}.\\
\indent In this work, we introduce CellStyle, a method designed to enhance the instance segmentation performance for cell microscopy images without requiring annotation labels for the target dataset.
CellStyle leverages existing annotated datasets to address the limitations of generalist zero-shot segmentation methods.
It achieves this by adapting key visual attributes of microscopy images—such as texture, color, and noise—from an unannotated target dataset to an annotated source dataset using a pretrained diffusion backbone.
This transformation preserves the original cell shapes, allowing the generated images to retain the annotations of the source dataset while ensuring they reflect the visual characteristics of the target dataset. 
\begin{figure}[tbp]
\centering
\includegraphics[width=0.9\textwidth]{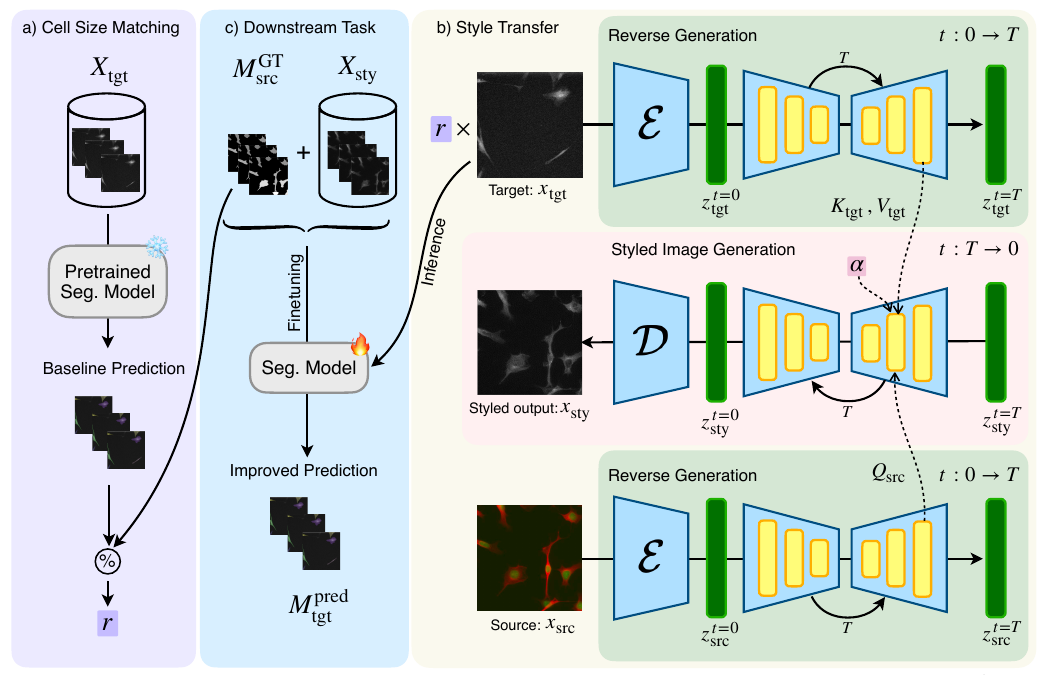}
\caption{Overview of the CellStyle pipeline: (a) Cell Size Matching: the average cell length in $X_{\mathrm{tgt}}$ is estimated using a pretrained segmentation model and compared to $X_{\mathrm{src}}$, to compute the cell size ratio $r$ which is used to scale $x_{\mathrm{tgt}}$; (b) Style Transfer: the pretrained diffusion model generates $x_{\mathrm{sty}}$ based on $x_{\mathrm{tgt}}$ and $x_{\mathrm{src}}$; (c) Downstream task: $X_{\mathrm{sty}}$ and ground truth labels $M^{\mathrm{GT}}_{\mathrm{src}}$ from $X_{\mathrm{tgt}}$ are used to finetune segmentation models.}
\label{fig:main}
\end{figure}
By finetuning pretrained segmentation models on style-transferred images paired with source annotations—without using any labels from the target dataset—we enable a form of zero-shot adaptation, where the model can be applied directly to the unannotated target data.
Our main contributions are threefold: (1) To the best of our knowledge, CellStyle is the first work proposing zero-shot domain transfer for cell microscopy imaging without requiring the training of a generative model. (2) We generate a diverse set of synthetic datasets that can be used to improve the downstream task performance on target data. (3) Experimental results demonstrate that CellStyle significantly enhances the zero-shot performance of cell segmentation models across various datasets.
\section{Method}
Diffusion models are generative models designed to synthesize data from pure noise by learning a data distribution \cite{ho2020denoising}.
Training involves progressively adding noise to a clean image sample and learning to predict this noise using the following loss function:
\begin{equation}
L(\theta)=\mathbb{E}_{x, \epsilon, t}\left[\left\|\epsilon-\epsilon_\theta\left(x_t, t\right)\right\|^2\right],
\end{equation}
where $\epsilon$ is the noise added to a clean image $x_0$ and $\epsilon_\theta(x_t, t)$ is the prediction for this noise based on $x_t$, $t$ and the learned parameters $\theta$.\\
\noindent The inference begins with a pure noise sample $x_T$ and a clean image is iteratively generated over $T$ steps.
Denoising Diffusion Implicit Models (DDIMs) follow the same training procedure but achieve significantly faster inference \cite{songdenoising}.
This is accomplished by formulating the diffusion process as a non-Markovian model, in contrast to the Markovian process used in \cite{ho2020denoising}.
As a result, during inference, DDIMs can generate images of comparable quality in approximately 50 iterations, rather than requiring $T \sim 10^3$ iterations.
Building on DDIMs, Latent Diffusion Models further improve inference speed by performing diffusion in a spatially compact latent space \cite{rombach2022high}.
An autoencoder \cite{esser2021taming} is used to encode the images into latent representations before the diffusion process and reconstruct them back into the image space after the diffusion process.
To enable style transfer on cell microscopy images without requiring additional training of a generative model on labeled data, we adopt Stable Diffusion (SD) as our diffusion backbone, which leverages this latent diffusion architecture.

To adapt an annotated source image $x_{\mathrm{src}}$ to a target image $x_{\mathrm{tgt}}$ with no annotations, we incorporate the finding outlined in \cite{hertzprompt,chung2024style}.
The finding shows that queries ($Q$) in the SD UNet attention blocks govern the shapes and spatial layouts of the generated objects, while keys ($K$) and values ($V$) control the other visual attributes such as texture, color, brightness, etc.
Specifically, after encoding the images $x_{\mathrm{src}}$ and $x_{\mathrm{tgt}}$ into the lower-dimensional representations $z^{t=0}_{\mathrm{src}}$ and $z^{t=0}_{\mathrm{tgt}}$ using the autoencoder from SD, we predict their corresponding noisy latent representations $z^{t=T}_{\mathrm{src}}$ and $z^{t=T}_{\mathrm{tgt}}$ by simulating a reverse generation process from $t=0$ to $t=T$ \cite{songdenoising}.
During this reverse process, we cache the queries ($Q_{\mathrm{src}}$) corresponding to $z^{t}_{\mathrm{src}}$ and keys and values corresponding to $z^{t}_{\mathrm{tgt}}$ ($K_{\mathrm{tgt}}$ and $V_{\mathrm{tgt}}$) at multiple levels of the self-attention blocks in the decoder of the SD UNet. 
Next, starting with $z^{t=T}_{\mathrm{sty}} = z^{t=T}_{\mathrm{src}}$, we generate the styled latent $z^{t=0}_{\mathrm{sty}}$ corresponding to $x_{\mathrm{sty}}$ using $Q_{\mathrm{src}}$, $K_{\mathrm{tgt}}$ and $V_{\mathrm{tgt}}$ (see Fig. \ref{fig:main}).
However, directly applying this approach to pairs of cell microscopy images can result in weak style transfer, depending on the disparity in average cell sizes between the datasets $X_{\mathrm{src}}$ and $X_{\mathrm{tgt}}$.
This issue arises because when computing \textit{Attention($Q_{\mathrm{src}}$, $K_{\mathrm{tgt}}$, $V_{\mathrm{tgt}}$)}, image patches from different images may lack strong feature correspondences when object sizes differ significantly.
To address this, we first compute the average cell lengths for both $X_{\mathrm{src}}$ and $X_{\mathrm{tgt}}$ and use the resulting cell size ratio $r$ to resize the images in $X_{\mathrm{tgt}}$ before performing style transfer.
The cell lengths for $X_{\mathrm{src}}$ are derived from the ground truth (GT) annotations $M^{\mathrm{GT}}_{\mathrm{src}}$, while a pretrained segmentation model \cite{mediar} is used to predict the annotations $M^{\mathrm{pred}}_{\mathrm{tgt}}$ for approximating the average cell lengths in $X_{\mathrm{tgt}}$ (see Fig. \ref{fig:main}a).
\newcolumntype{Y}{>{\centering\arraybackslash}p{2.0cm}}  
\newcolumntype{Z}{>{\centering\arraybackslash}p{0.7cm}}  
\newcolumntype{M}{>{\centering\arraybackslash}p{2.0cm}}  

\begin{table}[tbp]  
\centering
\caption{The selected pairs ($X_{\mathrm{src}}$, $X_{\mathrm{tgt}}$) for the experiments, the calculated average cell size ratios $r$, and the computed adaptive score scaling ratios $\alpha$.\\}
\begin{tabularx}{\textwidth}{Z@{\hspace{0pt}}c@{\hspace{0pt}}M@{\hspace{0pt}}Y@{\hspace{0pt}}Z@{\hspace{0pt}}Z | Z@{\hspace{0pt}}c@{\hspace{0pt}}M@{\hspace{0pt}}Y@{\hspace{0pt}}Z@{\hspace{0pt}}Z}
\toprule
 Pair && $X_{\mathrm{src}}$ & $X_{\mathrm{tgt}}$ & $r$ & $\alpha$ 
 & Pair && $X_{\mathrm{src}}$ & $X_{\mathrm{tgt}}$ & $r$ & $\alpha$ \\
\midrule
1 && MP6843 & Fluo-MSC & 1.0 & 1.5 
& 4 && Fluo-HeLa & Fluo-GOWT1 & 2.5 & 1.1 \\
2 && Huh7 & DIC-HeLa & 3.0 & 1.5 
& 5 && SHY5Y & MP6843 & 3.5 & 1.2 \\
3 && BV-2 & Fluo-HeLa & 2.1 & 1.2 
& 6 && NuI Kidney & NuI Cardia & 1.0  & 1.0 \\
\bottomrule
\end{tabularx}
\label{table:pairs}
\end{table}
\FloatBarrier   

\indent Another important consideration is the decrease in the attention map values obtained with ($Q_{\mathrm{src}}$, $K_{\mathrm{tgt}}$) compared to the original self-attention maps.
This is because the correspondence between $Q$ and $K$ is higher when they are derived from the same image.
To account for this, \cite{chung2024style} computes the average standard deviation ratio between the original self-attention scores and those generated using $Q$ and $K$ from different images. 
The attention scores are then scaled based on this ratio to maintain consistency.
However, when generating data across different dataset pairs with varying degrees of similarity, using a fixed scaling ratio for attention scores is not optimal (the corresponding experimental results are given in Section \ref{section:experiments}).
Instead, we propose computing this ratio separately for each dataset pair, which we term as \textit{adaptive score scaling ratio} $\alpha$.
Prior to performing style transfer, the standard deviations of the attention scores are computed across a small set of randomly selected samples from $X_{\mathrm{src}}$ and $X_{\mathrm{tgt}}$. 
These values are then averaged over all diffusion timesteps $T$ to calculate $\alpha$.
Subsequently, during style transfer, the attention maps computed between $Q_{\mathrm{src}}$ and $K_{\mathrm{tgt}}$ within the self-attention blocks of SD are scaled accordingly.
\section{Experiments}\label{section:experiments}
\textbf{Datasets:} We conduct experiments using publicly available datasets, including MP6843 from the Cell Image Library \cite{mp6843}; BV-2, Huh7, and SHSY5Y from LiveCell \cite{edlund2021livecell}; DIC-C2DH-HeLa (DIC-HeLa), Fluo-N2DL-HeLa (Fluo-HeLa), Fluo-C2DL-MSC (Fluo-MSC), and Fluo-N2DH-GOWT1 (Fluo-GOWT1) from the Cell Tracking Challenge (CTC) \cite{mavska2023cell}; and human kidney and cardia datasets from NuInsSeg \cite{Mahbod2024}.\\
\textbf{Experimental Setup:}
We conduct our experiments using the pretrained SD v1.5 model with 50 diffusion timesteps.
The experiments also include the use of different numbers of timesteps; however, due to space limitations, we cannot include those results in this paper.
For each dataset pair, we pick 4,000 combinations $\{(x_{\mathrm{src}}^i, x_{\mathrm{tgt}}^i)\}_{i=1}^{4000}$ from 
($X_{\mathrm{src}}$, $X_{\mathrm{tgt}}$) and generate the corresponding styled images $\{x_{\mathrm{sty}}^{i}\}_{i=1}^{4000}$ using an Nvidia L40S GPU.
During the generation process, we extract $Q_{\mathrm{src}}$, $K_{\mathrm{tgt}}$, and $V_{\mathrm{tgt}}$ from the last six attention layers of the model to be used when generating $X_{\mathrm{sty}}$.
To demonstrate the capabilities of CellStyle, we present experimental results on six dataset pairs (see Table \ref{table:pairs}).
The pairings are made based on the morphological characteristics of the cells in the images, with nuclei images paired with other nuclei images and cytoplasm images paired with other cytoplasm images.
Additionally, Table \ref{table:pairs} provides the predicted average cell size ratios ($r$) and the adaptive score scaling ratios ($\alpha$) computed for each dataset pair. 
\begin{figure}[tbp]
\centering
\includegraphics[width=0.9\textwidth]{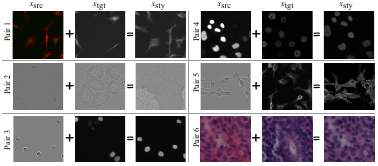}
\caption{Sample qualitative results ($x_{\mathrm{sty}}$) for each pair along with the corresponding source ($x_{\mathrm{src}}$) and the target ($x_{\mathrm{tgt}}$) images.}
\label{fig:qualitative}
\end{figure} \unskip \\
\noindent\textbf{Evaluation Methods:}
We evaluate CellStyle on the downstream task of instance segmentation using Cellpose \cite{cellpose}, Stardist \cite{stardist}, and Mediar \cite{mediar}.
First, we assess the performance of the pretrained models on $X_{\mathrm{tgt}}$ and then finetune them separately on $X_{\mathrm{src}}$ and $X_{\mathrm{sty}}$ using the models' default parameter configurations.
The training data for the base Cellpose model originally contains images from the MP6843 dataset.
For proper experimentation on pairs 1 and 5, we manually remove those images from the training dataset and train Cellpose from scratch.
This modified version of the base Cellpose model is used exclusively for experiments on these pairs, while the original base model is applied to all other pairs.
Stardist was originally trained on images featuring star-convex shapes, limiting its applicability to other cell shapes.
Therefore, we train it from scratch on the Cellpose training data for better generalization.
The training data for the pretrained Mediar contains a collection of datasets, including Cellpose \cite{cellpose}, DataScienceBowl 2018 \cite{caicedo2019nucleus}, LiveCell \cite{edlund2021livecell} and Omnipose \cite{cutler2022omnipose}.
Again, to ensure proper experimentation, we exclude the images used in our experiments from the Mediar training set and retrain the model from scratch.
This process is carried out separately for each pair, letting images from one pair be included in the training process for another pair.
We further compare our results to \cite{svoboda2016mitogen,bahr2021cellcyclegan,yilmaz2024cascaded,yilmaz2024annotated,eschweiler2024denoising} that also generate annotated synthetic cell microscopy data. 
However, they rely on labeled target data during generative model training, which presents a disadvantage for CellStyle.
To compensate for this in the comparisons to those models, we use a mixture of our synthetic data and a set of held-out target data with labels when finetuning the segmentation models. \\
\indent For the final quantitative evaluations, we use the segmentation accuracy measure (SEG) and the detection accuracy measure (DET) \cite{mavska2023cell}.
SEG measures how accurately the cell structures are spatially segmented while DET assesses whether the cells are correctly detected.
For false detections, DET introduces a penalty term that reduces the final score.
\newcolumntype{Y}{>{\centering\arraybackslash}X}
\newcolumntype{s}{>{\centering}p{2pt}}
\begin{table*}[tbp]
\centering
\caption{The quantitative results for Cellpose \cite{cellpose}, Stardist \cite{stardist}, and Mediar \cite{mediar} for each pair. For each segmentation model, the performance of the base model, the finetuned model on $X_{\mathrm{src}}$, and $X_{\mathrm{sty}}$ are given.\\}
\begin{tabularx}{\textwidth}{lcYYcYYcYYcYYcYYcYY}
\toprule
\multirow{2}[2]{*}{Method} && \multicolumn{2}{c}{Pair 1} && \multicolumn{2}{c}{Pair 2} && \multicolumn{2}{c}{Pair 3} && \multicolumn{2}{c}{Pair 4} && \multicolumn{2}{c}{Pair 5} && \multicolumn{2}{c}{Pair 6} \\
\cmidrule{3-4} \cmidrule{6-7} \cmidrule{9-10} \cmidrule{12-13} \cmidrule{15-16} \cmidrule{18-19} 
&&SEG & DET && SEG & DET && SEG & DET && SEG & DET && SEG & DET && SEG & DET \\
\midrule
Cellpose && 0.20 & 0.30 && 0.79 & 0.55 && 0.76 & 0.92 && \textbf{0.90} & 0.0 && 0.62 & 0.66 && 0.31 & 0.54  \\ 
Cellpose+$X_{\mathrm{src}}$ && 0.48 & 0.72 && 0.69 & 0.55 && 0.76 & 0.92 && 0.45 & 0.22 && 0.0 & 0.0 && 0.52 & 0.82  \\ 
Cellpose+$X_{\mathrm{sty}}$ && \textbf{0.62} & \textbf{0.83} && \textbf{0.81} & \textbf{0.85} && \textbf{0.81} & \textbf{0.95} && 0.79 & \textbf{0.93} && \textbf{0.76} & \textbf{0.89} && \textbf{0.55} & \textbf{0.83} \\ 
\midrule
Stardist && 0.0 & 0.08 && 0.26 & 0.21 && 0.75 & \textbf{0.95} && 0.69 & 0.94 && \textbf{0.36} & 0.46 && 0.10 & 0.18  \\ 
Stardist+$X_{\mathrm{src}}$ && 0.02 & 0.10 && 0.28 & 0.22 && 0.64 & 0.84 && 0.41 & 0.82 && 0.01 & 0.0 && 0.52 & 0.80  \\ 
Stardist+$X_{\mathrm{sty}}$ && \textbf{0.28} & \textbf{0.56} && \textbf{0.60} & \textbf{0.77} && \textbf{0.77} & 0.94 && \textbf{0.87} & \textbf{0.98} && 0.32 & \textbf{0.47} && \textbf{0.56} & \textbf{0.81}  \\ 
\midrule
Mediar && 0.26 & 0.58 && 0.83 & 0.96 && 0.66 & 0.95 && 0.85 & 0.94 && 0.67 & 0.85 && 0.48 & 0.71  \\ 
Mediar+$X_{\mathrm{src}}$ && 0.24 & \textbf{0.59} && 0.83 & 0.96 && 0.68 & 0.96 && 0.77 & 0.95 && 0.66 & 0.84 && 0.57 & 0.87  \\ 
Mediar+$X_{\mathrm{sty}}$ && \textbf{0.31} & 0.54 && \textbf{0.85} & \textbf{0.97} && \textbf{0.77} & \textbf{0.97} && \textbf{0.87} & \textbf{0.98} && \textbf{0.69} & \textbf{0.90} && \textbf{0.58} & \textbf{0.89}  \\ 
\bottomrule
\end{tabularx}
\label{table:main}
\end{table*}
\newcolumntype{Y}{>{\centering\arraybackslash}X}

\begin{table}[tbp]
\centering
\caption{Segmentation results for (a) Fluo-GOWT1 and (b) Fluo-HeLa on Cellpose \cite{cellpose}, Stardist \cite{stardist}, and Mediar \cite{mediar}. For fairness, the CellStyle outputs are combined with real target images to finetune the segmentation models, as other methods use labeled target data when training the generative models. The best-performing results are highlighted in bold, while the second-best are underlined.}
\label{table:others}
\begin{subtable}{\textwidth}
    \centering
    \caption{Comparative segmentation results for Fluo-GOWT1}
    \label{table:gowt}
    \begin{tabularx}{\textwidth}{lcYYcYYcYY}
        \toprule
        \multirow{2}[2]{*}{Method} && \multicolumn{2}{c}{\cite{eschweiler2024denoising}} 
        && \multicolumn{2}{c}{\cite{yilmaz2024cascaded}} 
        && \multicolumn{2}{c}{Ours+Real} \\
        \cmidrule{3-4} \cmidrule{6-7} \cmidrule{9-10} 
        && SEG & DET && SEG & DET && SEG & DET \\
        \midrule
        Cellpose && \underline{0.88} & 0.91 && 0.87 & \underline{0.96} && \textbf{0.91} & \textbf{0.98} \\
        Stardist && 0.44 & 0.79 && \underline{0.84} & \underline{0.86} && \textbf{0.87} & \textbf{0.98} \\
        Mediar && 0.79 & \underline{0.78} && \underline{0.91} & \textbf{0.97} && \textbf{0.92} & \textbf{0.97} \\
        \bottomrule
    \end{tabularx}
\end{subtable}


\begin{subtable}{\textwidth}
    \centering
    \caption{Comparative segmentation results for  Fluo-HeLa}
    \label{table:fluon2dl}
    \begin{tabularx}{\textwidth}{lcYYcYYcYYcYYcYYcYY}
        \toprule
        \multirow{2}[2]{*}{Method} 
        && \multicolumn{2}{c}{\cite{svoboda2016mitogen}} 
        && \multicolumn{2}{c}{\cite{eschweiler2024denoising}} 
        && \multicolumn{2}{c}{\cite{yilmaz2024cascaded}} 
        && \multicolumn{2}{c}{\cite{yilmaz2024annotated}} 
        && \multicolumn{2}{c}{\cite{bahr2021cellcyclegan}} 
        && \multicolumn{2}{c}{Ours+Real} \\
        \cmidrule{3-4} \cmidrule{6-7} \cmidrule{9-10} \cmidrule{12-13} \cmidrule{15-16} \cmidrule{18-19} 
        && SEG & DET && SEG & DET && SEG & DET && SEG & DET && SEG & DET && SEG & DET \\
        \midrule
        Cellpose && 0.70 & 0.88 && 0.75 & \underline{0.94} && \underline{0.80} & 0.93 && 0.71 & 0.83 && 0.76 & 0.81 && \textbf{0.83} & \textbf{0.96} \\
        Stardist && 0.70 & \textbf{0.98} && 0.62 & 0.87 && 0.72 & 0.91 && \underline{0.75} & 0.85 && \underline{0.75} & \underline{0.95} && \textbf{0.77} & \underline{0.95} \\
        Mediar && 0.83 & \underline{0.97} && 0.68 & 0.88 && 0.84 & \underline{0.97} && 0.81 & \textbf{0.98} && \underline{0.85} & \underline{0.97} && \textbf{0.88} & \textbf{0.98} \\
        \bottomrule
    \end{tabularx}
\end{subtable}

\end{table}
\noindent\textbf{Experimental Results:}
We present our experimental results in a zero-shot setting for the selected segmentation models in Table \ref{table:main}.
The first row for each model reports the SEG and DET scores for the generalizable pretrained base models.
Next, we provide the scores obtained when the base model is finetuned on $X_{\mathrm{src}}$ and evaluated on $X_{\mathrm{tgt}}$.
Finally, the results when the base model is finetuned using {$\{X_{\mathrm{sty}}$, $M^{\mathrm{GT}}_{\mathrm{src}}\}$} are presented.
In general, the base Stardist model tends to underperform compared to other methods.
This limitation is inherent to its design, as it was specifically developed for segmenting star-convex shapes, such as cell nuclei.
As a result, its performance declines on non-convex cytoplasm images, particularly on pairs 1 and 5.
As shown in Table \ref{table:main}, CellStyle significantly improves the zero-shot segmentation quality compared to the base models in terms of both SEG and DET.
It is important to note that the overall segmentation performance should be assessed by considering the average of SEG and DET scores \cite{mavska2023cell}.
This is mainly because the GT segmentation labels are not present for all the cells in the CTC dataset, while cell centers are fully annotated.
Since SEG is computed based on these GT masks, only the regions containing annotations are evaluated, meaning that false positives in the predicted masks are not penalized.
However, this limitation can be overcome by considering the arithmetic mean of SEG and DET, referred to as the overall performance measure (OP\textsubscript{CSB}) \cite{mavska2023cell}.
For example, in Table \ref{table:main}, the base Cellpose model achieves a higher SEG score for Pair 4, while upon inspecting the predicted masks, we observed numerous background regions that were falsely segmented as cells.
This cannot be captured directly by SEG, hindering proper comparisons of the predictions.
Alternatively, when OP\textsubscript{CSB} is considered, the performance of our approach (with 0.5$\times$(SEG + DET) $=0.86$) significantly surpasses the base model (with 0.5$\times$(SEG + DET)  $=0.45$) even for this particular pair.
Additionally, in Table \ref{table:others}, we compare our method to other generative models on the downstream segmentation task in a non-zero-shot setting since they use labeled data during the training of generative models.
The comparisons are conducted using the datasets that were also generated by these methods, namely Fluo-HeLa and Fluo-GOWT1.
The compared methods were specifically designed for these datasets, and adapting them to all the datasets used in this work was not feasible without major modifications.
While our approach demonstrates competitive performance to those models even for the zero-shot scenario (see Table \ref{table:main}), when combined with real data, it surpasses them at least by 1\% in terms of OP\textsubscript{CSB}, the average of SEG and DET (see Table \ref{table:others}).
\newcolumntype{Y}{>{\centering\arraybackslash}X}
\newcolumntype{s}{>{\centering}p{2pt}}

\begin{table}[tbp]
\centering
\caption{The results for the ablation experiments using Cellpose \cite{cellpose}, Stardist \cite{stardist}, and Mediar \cite{mediar}. For the whole CellStyle pipeline denoted as $+X_{\mathrm{sty}}$, we calculated $\alpha=1.5$ or $r=1.0$ for some pairs (see Table \ref{table:pairs}). This causes the results for the configurations $+X_{\mathrm{sty}}-\alpha$ or $+X_{\mathrm{sty}}-r$ to coincide with the setting from $+X_{\mathrm{sty}}$ for those pairs. These results are replaced by '-' to avoid redundancy.\\}
\begin{tabularx}{\textwidth}{lcYYcYYcYYcYYcYYcYY}
\toprule
\multirow{2}[2]{*}{Method} && \multicolumn{2}{c}{Pair 1} && \multicolumn{2}{c}{Pair 2} && \multicolumn{2}{c}{Pair 3} && \multicolumn{2}{c}{Pair 4} && \multicolumn{2}{c}{Pair 5} && \multicolumn{2}{c}{Pair 6} \\
\cmidrule{3-4} \cmidrule{6-7} \cmidrule{9-10} \cmidrule{12-13} \cmidrule{15-16} \cmidrule{18-19} 
&&SEG & DET && SEG & DET && SEG & DET && SEG & DET && SEG & DET && SEG & DET \\
\midrule
Cellp+$r$ && 0.48 & 0.72 && 0.61 & 0.80 && 0.65 & 0.84 && 0.78 & 0.0 && 0.60 & 0.73 && 0.52 & 0.82  \\ 
Cellp+$X_{\mathrm{sty}}-\alpha$ && - & - && - & - && 0.80 & 0.93 && \textbf{0.79} & 0.90 && 0.72 & 0.81 && 0.54 & 0.81 \\ 
Cellp+$X_{\mathrm{sty}}-r$ && - & - && 0.43 & 0.0 && 0.59 & 0.82 && 0.39 & 0.63 && 0.04 & 0.0 && - & - \\
Cellp+$X_{\mathrm{sty}}$ && \textbf{0.62} & \textbf{0.83} && \textbf{0.81} & \textbf{0.85} && \textbf{0.81} & \textbf{0.95} && \textbf{0.79} & \textbf{0.93} && \textbf{0.76} & \textbf{0.89} && \textbf{0.55} & \textbf{0.83}  \\
\midrule
Strd+$r$ && 0.02 & 0.10 && 0.03 & 0.05 && 0.72 & 0.90 && 0.85 & 0.96 && 0.03 & 0.0 && 0.52 & 0.80  \\ 
Strd+$X_{\mathrm{sty}}-\alpha$ && - & - && - & - && 0.76 & \textbf{0.94} && 0.85 & \textbf{0.98} && 0.31 & 0.46 && 0.54 & 0.80 \\ 
Strd+$X_{\mathrm{sty}}-r$ && - & - && 0.52 & 0.73 && 0.18 & 0.30 && 0.39 & 0.67 && 0.03 & 0.0 && - & - \\
Strd+$X_{\mathrm{sty}}$ && \textbf{0.28} & \textbf{0.56} && \textbf{0.60} & \textbf{0.77} && \textbf{0.77} & \textbf{0.94} && \textbf{0.87} & \textbf{0.98} && \textbf{0.32} & \textbf{0.47} && \textbf{0.56} & \textbf{0.81}  \\
\midrule
Mdr+$r$ && 0.24 & 0.50 && 0.79 & 0.96 && 0.75 & 0.95 && 0.85 & 0.95 && 0.63 & 0.84 && 0.57 & 0.87  \\ 
Mdr+$X_{\mathrm{sty}}-\alpha$ && - & - && - & - && 0.68 & 0.96 && \textbf{0.87} & \textbf{0.98} && 0.68 & 0.89 && 0.56 & 0.88  \\ 
Mdr+$X_{\mathrm{sty}}-r$ && - & - && 0.77 & 0.86 && 0.65 & 0.92 && 0.80 & 0.96 && 0.60 & 0.77 && - & - \\
Mdr+$X_{\mathrm{sty}}$ && \textbf{0.31} & \textbf{0.54} && \textbf{0.85} & \textbf{0.97} && \textbf{0.77} & \textbf{0.97} && \textbf{0.87} & \textbf{0.98} && \textbf{0.69} & \textbf{0.90} && \textbf{0.58} & \textbf{0.89}  \\
\bottomrule
\end{tabularx}
\label{table:ablations}
\end{table}
\noindent\textbf{Ablation Experiments:} To assess the significance of individual components in CellStyle, we conduct ablation experiments. Specifically, we test the effects of (i) performing only cell size matching and finetuning the segmentation models on $X_{\mathrm{src}}$, (ii) using a constant attention score scaling ratio of $\alpha=1.5$ during style transfer, (iii) performing style transfer without cell size matching, i.e., with $r=1.0$, and (iv) using the whole CellStyle pipeline (see Table \ref{table:ablations}).
When only cell size matching is used ($+r$), the performance tends to drop, which is more pronounced for the pairs that are significantly different in terms of structure or color. 
Similarly, when a fixed attention score scaling ratio $\alpha=1.5$ is used for all the pairs ($+X_{\mathrm{sty}}-\alpha$), the overall segmentation performance decreases compared to the adaptive approach, where $\alpha$ is specifically calculated for each pair of datasets.
Intuitively, the value of this parameter varies based on the similarity of cell characteristics between the paired datasets, with higher values of $\alpha$ for lower similarity, and vice versa.
Additionally, performing style transfer without cell size matching ($+X_{\mathrm{sty}}-r$) results in a reduction in segmentation quality primarily due to structural size differences between $X_{\mathrm{src}}$ and $X_{\mathrm{tgt}}$ causing blurry or indistinct representations in $X_{\mathrm{sty}}$.
\section{Conclusion}
We introduced CellStyle to improve the zero-shot performance of pretrained cell instance segmentation models.
By leveraging style transfer, CellStyle transforms a labeled source dataset to match the visual characteristics of an unlabeled target dataset while preserving cell morphology.
The generated styled images, combined with the source annotations, enable the finetuning of segmentation models without requiring additional target dataset labels.
Our experimental results demonstrate that CellStyle significantly enhances segmentation performance compared to baseline methods and alternative generative models.
\noindent\textbf{Acknowledgments:} This work was partially funded by the German Research Foundation DFG (STE2802/5-1).

\subsubsection{\discintname}
The authors have no competing interests to declare that are relevant to the content of this article.

\bibliographystyle{splncs04}
\bibliography{Paper-1040}
\end{document}